\documentclass[conference]{IEEEtran}
\IEEEoverridecommandlockouts
\usepackage{cite}
\usepackage{amsmath,amssymb,amsfonts}
\usepackage{algorithmicx}
\usepackage{graphicx}
\usepackage{textcomp}
\usepackage[table]{xcolor}

\usepackage{bm}			
\usepackage{soul} 
\usepackage{algpseudocode}

\usepackage{hyperref}
\usepackage[a4paper, margin=3cm]{geometry}

\usepackage{blindtext}
\usepackage{chngpage}
\usepackage{parskip}
\usepackage{setspace}

\if CLASSOPTION compsoc
\usepackage[caption=false,font=normalsize,labelfont=sf,textfont=sf]{subfig}
\else
\usepackage[caption=false,font=footnotesize]{subfig}
\fi

\algrenewcommand\algorithmicindent{0.6em}%

\def\BibTeX{{\rm B\kern-.05em{\sc i\kern-.025em b}\kern-.08em
    T\kern-.1667em\lower.7ex\hbox{E}\kern-.125emX}}
\begin{document}

\title{Distilling Deep RL Models Into Interpretable Neuro-Fuzzy Systems}

\author{
	\IEEEauthorblockN{Arne Gevaert, Jonathan Peck and Yvan Saeys}
	\IEEEauthorblockA{
		Department of Applied Mathematics and Statistics\\
		Ghent University, Ghent, Belgium\\
		Email: \{arne.gevaert, jonathan.peck, yvan.saeys\}@ugent.be
	}
}

\maketitle
\begin{abstract}
Deep Reinforcement Learning uses a deep neural network to encode a policy, which achieves very good performance in a wide range of applications but is widely regarded as a black box model. A more interpretable alternative to deep networks is given by neuro-fuzzy controllers. Unfortunately, neuro-fuzzy controllers often need a large number of rules to solve relatively simple tasks, making them difficult to interpret. In this work, we present an algorithm to distill the policy from a deep Q-network into a compact neuro-fuzzy controller. This allows us to train compact neuro-fuzzy controllers through distillation to solve tasks that they are unable to solve directly, combining the flexibility of deep reinforcement learning and the interpretability of compact rule bases. We demonstrate the algorithm on three well-known environments from OpenAI Gym, where we nearly match the performance of a DQN agent using only 2 to 6 fuzzy rules.
\end{abstract}

\begin{IEEEkeywords}
Neuro-fuzzy, Distillation, DQN, Reinforcement Learning.
\end{IEEEkeywords}

\section{Introduction}
Recently, significant progress has been made in the field of Deep Reinforcement Learning, with advances in a wide variety of application domains, such as arcade game playing \cite{Mnih2013}, continuous control \cite{Lillicrap2016}, and beating professional players in the game of Go \cite{Silver2017}. Most of these advances have been made using deep neural networks, which are widely regarded as \textit{black boxes} \cite{Frosst2017, Montavon2017}. This means that the inner workings of a deep neural network are difficult to understand, making interpretation of the learned policy by humans a difficult task. However, interpretability is a critical property of machine learning models in many application domains, including legal and medical applications \cite{Zech2018}. Recent EU legislation even requires many machine learning models to be interpretable by default \cite{VOIGT2018}.

An alternative to deep neural networks is the use of fuzzy controllers. These controllers consist of a set of fuzzy IF-THEN rules (e.g. ``IF temp IS low THEN output IS high''), which can make them much more interpretable than neural networks, provided they are relatively compact. Fuzzy controllers can either be constructed manually using expert knowledge \cite{MartinLarsen1980} or they can be learned from data \cite{Jang1993}. A popular type of fuzzy controller that can be learned from data is the \textit{neuro-}fuzzy controller \cite{jang1995}. In this case, the calculations are organised in a neural network-like structure. The gradient of the output is then calculated using backpropagation and the parameters are optimized using gradient descent. Neuro-fuzzy controllers have successfully been applied in a wide variety of domains, including nonlinear modeling of dynamic systems \cite{bartczuk2016}, control of robotic systems \cite{chatterjee2005} and power systems \cite{shihabudheen2018}.

A possible approach to improve the interpretability of deep reinforcement learning algorithms is to replace the deep neural network with a neuro-fuzzy controller. In this work, we apply this approach to the Deep Q-Network (DQN) algorithm \cite{Mnih2013}. Unfortunately, as will be shown in the results in Section \ref{sec:results}, simply replacing the neural network in the DQN algorithm with a neuro-fuzzy controller does not yield satisfying results on even the simplest environments. We therefore propose a policy distillation algorithm \cite{Rusu2015} to distill the knowledge from a trained deep Q-network into a compact neuro-fuzzy controller. We extend the original policy distillation algorithm with a pre-processing and post-processing step and introduce specific regularization terms to maximize model interpretability. We show how this approach allows us to train compact neuro-fuzzy controllers to solve tasks that they are unable to solve by direct reinforcement learning, combining the flexibility of deep Q-learning with the interpretability of compact neuro-fuzzy controllers.

We first give an overview of related work in section \ref{sec:related_work}. Next, we explore the DQN algorithm in section \ref{sec:dqn} and policy distillation in section \ref{sec:policy_distillation}. In section \ref{sec:fuzzy_sets_neuro_fuzzy_control}, we describe the mechanisms behind the neuro-fuzzy controller used in this work. The complete algorithm is described in detail in section \ref{sec:methods}. We then apply the algorithm to a number of problems in section \ref{sec:results} and conclude in section \ref{sec:discussion}.

\section{Related Work}
\label{sec:related_work}
The technique of knowledge distillation has originally been applied to compress an ensemble of models into a single neural network for classification \cite{Hinton2015}. This greatly reduces the computational cost of inference while retaining the superior performance of the ensemble model.
The extension of this technique to the domain of reinforcement learning is called \textit{policy distillation} \cite{Rusu2015}. Not only can the size of a deep Q-network (DQN) greatly be reduced using policy distillation, but it is also possible to merge several task-specific networks into a single, more general DQN.

Although the original purpose of distillation is compression and generalization, it can also be applied to improve interpretability of models. In this case, the knowledge from a (large) model is distilled into a more interpretable representation, such as decision trees \cite{Liu2019} or soft decision trees \cite{Frosst2017}.
This same reasoning has also been applied to reinforcement learning and policy distillation \cite{coppens}. The knowledge encoded by a deep Q-network or an explicit policy network can analogously be distilled into a representation that is considered more interpretable. In this work, we use neuro-fuzzy controllers as a target model to distill the knowledge represented by a deep Q-network. A recent overview of techniques to increase the explainability of reinforcement learning algorithms is given by \cite{puiutta2020}.

\section{The DQN Algorithm}
\label{sec:dqn}
In (deep) reinforcement learning, an \textit{agent} interacts with an \textit{environment} by taking \textit{actions} in that environment. When the agent takes an action $a \in \mathcal{A}$, the environment responds with a resulting \textit{state} $s \in \mathcal{S}$ and a \textit{reward} $r \in \mathbb{R}$. The agent then takes a new action $a'$, to which the environment responds with a new state $s'$ and reward $r'$. This cycle continues until some end state or a maximum number of iterations is reached. The goal of reinforcement learning is to maximize the total reward $r$ over some time period.

DQN aims to train a policy that maximizes the \textit{return} $R = \sum_{t=0}^\infty \gamma^{t}r_t$, where $\gamma$ is a constant between 0 and 1 that controls the importance that the agent assigns to rewards that are further in the future. The deep neural network in DQN is trained to approximate the optimal action-value function $Q^\ast : \mathcal{S} \times \mathcal{A} \rightarrow \mathbb{R}$ that tells us what the return would be, if a given action $a$ is taken in state $s$, and an optimal policy is followed afterwards. This function can be used to construct the optimal policy: $\pi^\ast(s) = \arg\max_a Q^\ast(s,a)$.

The algorithm works by interacting with the environment and saving transitions $(s,a,s',r)$ into a replay memory $D$. After each step, a batch of transitions $B$ is sampled from $D$, and the neural network is trained to minimize the loss function $\mathcal{L} = (Q(s,a) - (r + \gamma \max_{a'}(Q(s', a'))))^2$. This loss function is derived from the \textit{Bellman equation}, and ensures that the network will converge to the optimal action-value function $Q^\ast$ \cite{Mnih2013}.

\section{Policy Distillation}
\label{sec:policy_distillation}
Distillation is a method to transfer knowledge from a \textit{teacher} model $T$ to a \textit{student} model $S$. In supervised learning, this is done by using the output distributions of the teacher to train the student \cite{Hinton2015}. As these output distributions (\textit{soft targets}) contain confidence levels of the teacher over all output classes, they contain more information than the original labels (\textit{hard targets}), allowing the student model to learn more efficiently. However, the output distribution for a fully trained teacher model usually has the correct class at very high probability, with all the other classes very close to 0. As such, the soft target hardly provides any information beyond the ground truth labels encoded in the original hard targets. For this reason, the original softmax output function $\sigma(\bm{z})$ of the teacher network is replaced with a \textit{temperature softmax} function $\sigma_\tau(\bm{z})$ that converts each logit, $z_i$, computed for each class into a probability, $q_i$, by comparing $z_i$ with the other logits:
\begin{equation}
q_i = \sigma_\tau(\bm{z})_i = \frac{\exp(z_i/\tau)}{\sum_j \exp(z_j/\tau)}
\end{equation}
Note that, if the temperature $\tau=1$, we obtain the original softmax function. If $\tau>1$, the result of this function is a smoothed version of the normal softmax. The student is then trained on these smoothened distributions using the classical cross-entropy loss function.

\begin{figure}
	\centering
	\subfloat[Teacher trajectories]{\includegraphics[width=0.5\linewidth]{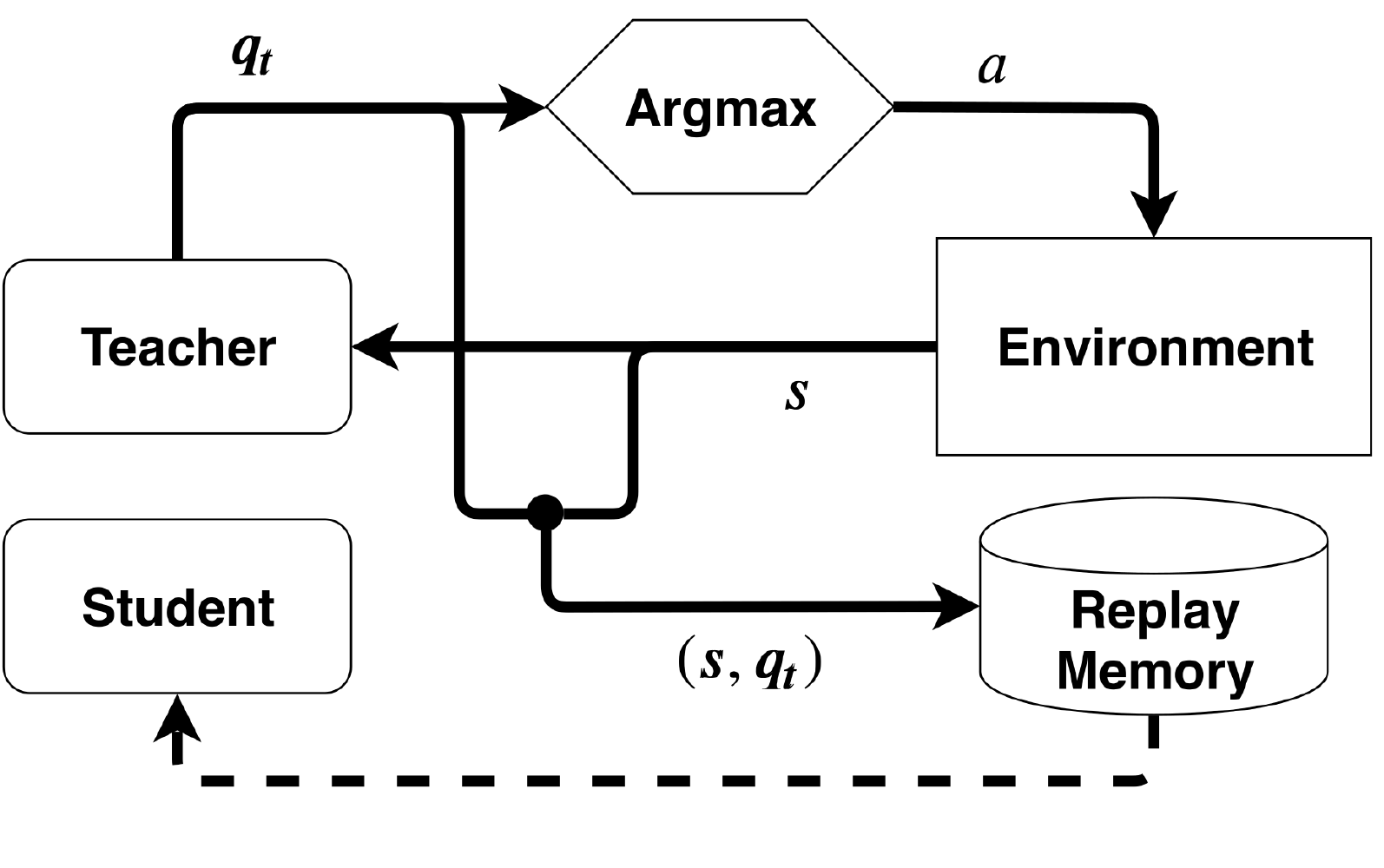}
		\label{fig:policy-distillation:teacher}}
	\subfloat[Student trajectories]{\includegraphics[width=0.5\linewidth]{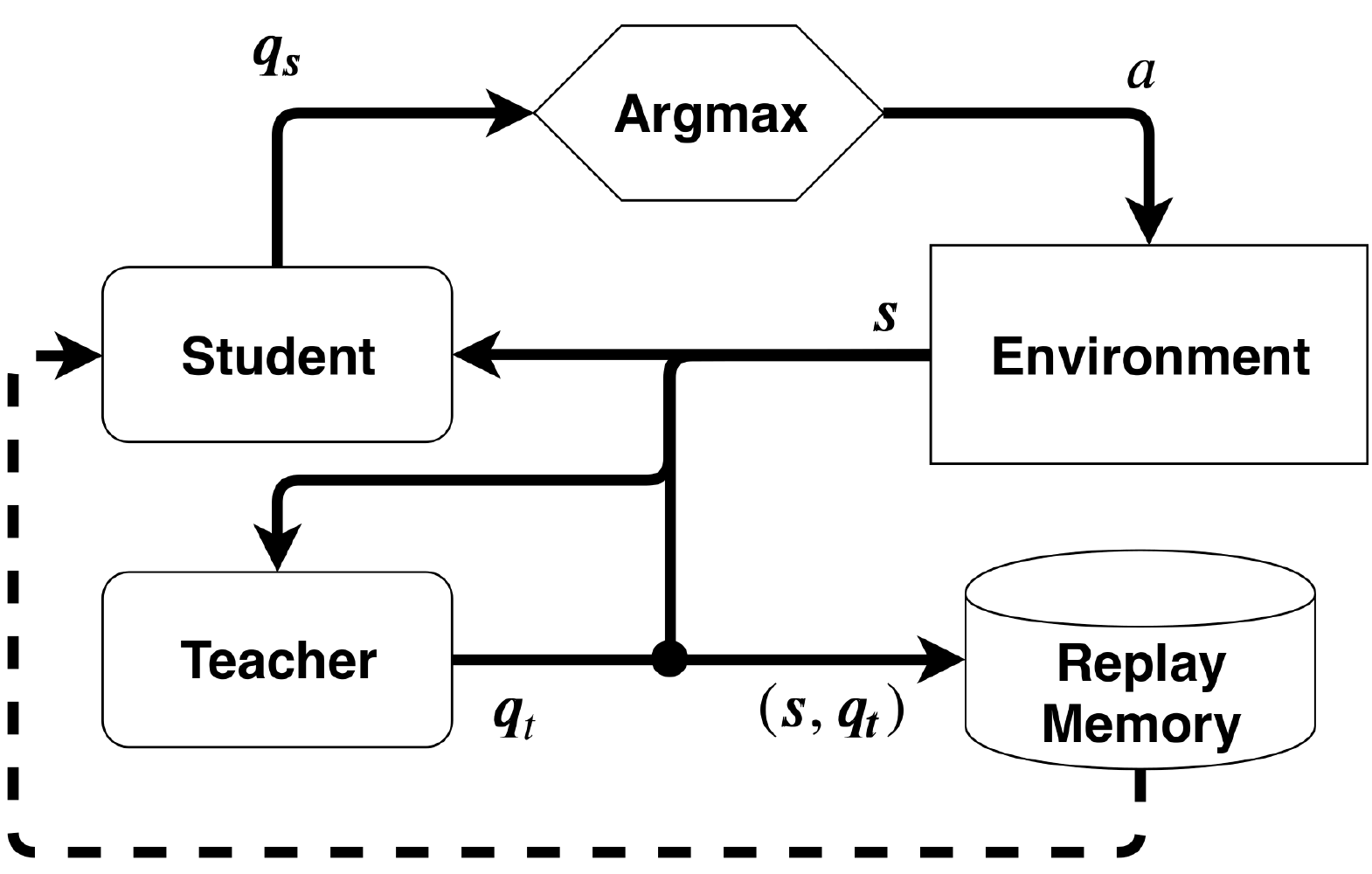}
		\label{fig:policy-distillation:student}}
	\caption{Comparison between policy distillation using teacher trajectories and policy distillation using student trajectories. In this work, the approach in Figure \ref{fig:policy-distillation:student} is used.}
	\label{fig:policy-distillation}
	\vspace{-1em}
\end{figure}

Policy distillation is the application of this idea to reinforcement learning agents. In \cite{Rusu2015}, a dataset is built by having the teacher model interact with the environment and recording the encountered states and output Q-values. The student is then trained to mimic this behaviour (see Figure \ref{fig:policy-distillation:teacher}). As we require the student to output Q-values, which do not form valid probability distributions, we replace the cross-entropy loss with a \textit{temperature Kullback-Leibler divergence} loss:
\begin{equation}
\label{eqn:temp-kl-div}
\mathcal{L}_{KL}(\bm{q}^T,\bm{q}^S) = \sigma_\tau(\bm{q}^T) \ln \frac{\sigma_\tau(\bm{q}^T)}{\sigma(\bm{q}^S)}
\end{equation}
Where $\bm{q}^T$ is the output of the teacher model and $\bm{q}^S$ the output of the student model. Note that the output of a softmax function on a vector of Q-values is usually a very smooth probability distribution, as Q-values often have very small yet important differences. For this reason, the temperature $\tau$ in policy distillation is usually lower than 1, resulting in sharper probability distributions.

More recent advances in policy distillation have shown better empirical results if an $\epsilon$-greedy student policy is followed during distillation \cite{Czarnecki, Schmitt2018, Parisotto2015}. This means that, for some small $\epsilon$ (e.g. $0.05$), a random action is chosen with probability $\epsilon$ and the student's action is chosen with probability $1-\epsilon$. The student is then trained on the output of the teacher for visited states, rather than building a dataset by having the teacher interact with the environment (see Figure \ref{fig:policy-distillation:student}). A possible explanation for this performance improvement is that the teacher policy is deterministic, while the student policy changes as the student is trained. This leads to a better exploration of the environment if student trajectories are followed. A schematic overview of both approaches is shown in Figure \ref{fig:policy-distillation}.

\section{Fuzzy Sets and Neuro-Fuzzy Control}
\label{sec:fuzzy_sets_neuro_fuzzy_control}
Fuzzy sets form an extension of the classical notion of a mathematical set \cite{Zadeh1975}. In a classical set, every element is either a member or not a member of the set. For a fuzzy set, the membership of an element in the set can take any value between 0 (not a member) and 1 (entirely a member). A fuzzy set $A$ in $\mathbb{R}$ is therefore often written as a \textit{membership function} $A: \mathbb{R} \rightarrow [0,1]$, mapping an input $x$ to its membership degree.

Neuro-fuzzy control applies this notion of fuzzy sets to create control systems, by defining a number of fuzzy IF-THEN rules and composing them to create a function approximator \cite{Jang1993}:
\begin{figure}[h]
	\centering
	\begin{tabular}{c}
		IF $x_1$ is $A_{11}$ AND $\dots$ AND $x_m$ is $A_{m1}$ \\ THEN $\bm{y}$ is $f_1(x_1,\dots,x_m)$\\
		IF $x_1$ is $A_{12}$ AND $\dots$ AND $x_m$ is $A_{m2}$ \\ THEN $\bm{y}$ is $f_2(x_1,\dots,x_m)$\\
		$\cdots$ \\
		IF $x_1$ is $A_{1n}$ AND $\dots$ AND $x_m$ is $A_{mn}$ \\ THEN $\bm{y}$ is $f_n(x_1,\dots,x_m)$\\
	\end{tabular}
\end{figure}

Where $n$ is the number of fuzzy rules, $m$ is the number of input dimensions, $\bm{x} = (x_1, ..., x_m)$ is the input vector, $\bm{y}$ is the output vector, each $A_{ij}$ is a fuzzy set in $\mathbb{R}$ (for input dimension $i$ and fuzzy rule $j$), and each $f_j$ is differentiable w.r.t. its inputs (in practice, linear or constant functions are used). The degree of activation of the $j$-th fuzzy rule is calculated as a triangular norm (T-norm) \cite{Klement2000} of the membership degrees of each input $x_i$ in the corresponding set $A_{ij}$. A T-norm is the fuzzy extension of a boolean AND-operation, in the form of a two-argument function $T: \mathbb{R}^2 \rightarrow [0,1]$. Any T-norm is defined to be associative, and can therefore be extended to any number of arguments by repeated application: $T(x_1, x_2, ..., x_n) = T(x_1, T(x_2, \dots, T(x_{n-1}, x_n) \dots)$ In most cases, the product T-norm and Gaussian membership functions are used. The degrees of activation are then calculated as:
\begin{align}
A_{ij}(x_i) &= \exp(-\frac{x_i - \mu_{ij}}{\sigma_{ij}})\label{eqn:gaussian_membership}\\
R_j &= \prod_{i=1}^{m} A_{ij}(x_i)\label{eqn:rule_activation}
\end{align}
Where $\mu_{ij}$ and $\sigma_{ij}$ are learned parameters representing the center and standard deviation of the Gaussian membership function for input dimension $i$ and fuzzy rule $j$. Inference is then performed by calculating the activation for each rule using (\ref{eqn:rule_activation}), normalizing these activations to sum to 1, and calculating the output as a weighted sum of the output functions using the activations as weights:
\begin{align}
\bar{R_i} &= \frac{R_i}{\sum_{j=1}^{n}R_j}\label{eqn:act_degree_norm}\\
\hat{\bm{y}} &= \sum_{i=1}^{n} \bar{R_i} f_i(x_1,\dots,x_m)\label{eqn:anfis_output}
\end{align}
Since the operations in (\ref{eqn:gaussian_membership}), (\ref{eqn:rule_activation}), (\ref{eqn:act_degree_norm}) and (\ref{eqn:anfis_output}) are all differentiable w.r.t. the parameters $\{\mu_{ij}\}$, $\{\sigma_{ij}\}$ and the parameters of $\{f_j\}$, backpropagation and gradient descent can be used to optimize these parameters. In other words, the system can be formulated and trained like a type of neural network, hence the name \textit{neuro-fuzzy} controller.

\section{Methods}
\label{sec:methods}
We first discuss the details of the neuro-fuzzy architecture used in the distillation process in section \ref{sec:neuro_fuzzy_control_using_weighted_tnorms}. The complete algorithm consists of three steps: pre-processing, distillation and post-processing. We discuss these steps in sections \ref{sec:pre-processing}, \ref{sec:distillation} and \ref{sec:post-processing} respectively. Finally, we discuss some regularization techniques used to improve the interpretability of the resulting neuro-fuzzy controller in section \ref{sec:regularization}.

\subsection{Neuro-Fuzzy Control using Weighted T-Norms}
\label{sec:neuro_fuzzy_control_using_weighted_tnorms}
We extend the neuro-fuzzy controller from section \ref{sec:fuzzy_sets_neuro_fuzzy_control} using a weighted T-norm \cite{Yager2004}. This is an extension of a normal T-norm, where an extra weight $w_i$ is associated with each argument: $T(x_1, x_2; w_1, w_2)$. This weight expresses the importance of the corresponding argument. If all weights $w_i$ are 1, we retrieve the original T-norm: $T(x_1, x_2; 1, 1) = T(x_1, x_n)$. If $w_i = 0$, then the argument $x_i$ has no influence on the output of the function. The weighted extension of the product T-norm is calculated as follows:
\begin{equation}
T_P(x_1, \dots, x_n; w_1, \dots, w_n) = \prod_{i=1}^{n} x_i^{w_i}
\end{equation}
This extension is incorporated into the model from section \ref{sec:fuzzy_sets_neuro_fuzzy_control} by changing (\ref{eqn:rule_activation}) into the following:
\begin{equation}
R_j = \prod_{i=1}^{m} A_{ij}(x_i)^{w_{ij}}
\label{eqn:weighted_tnorm}
\end{equation}
and adding the parameters $\{w_{ij}\}$ to the set of trainable parameters for the model. We also restrict the values for $w_{ij}$ to be non-negative and divide weights $w_{ij}$ by the maximal weight for their corresponding rule $\max_i \{w_{ij}\}$ before applying (\ref{eqn:weighted_tnorm}). This ensures that the most important fuzzy set in the antecedent of a rule always has an importance value of 1.

\subsection{Pre-Processing}
\label{sec:pre-processing}
Clustering algorithms can be used to find a good initialization for neuro-fuzzy controllers \cite{Paiva2004}. Given a dataset of $m$-dimensional input vectors $I$ and corresponding $n$-dimensional output vectors $U$, any clustering algorithm that outputs a set of centroid points (e.g. K-means) can be used to initialize a neuro-fuzzy controller as follows:
\begin{enumerate}
	\item Construct the matrix $M$ by concatenating every input vector $\bm{v}_i \in I$ to its corresponding output vector $\bm{u}_i \in U$.
	\item Apply a clustering algorithm to the matrix $M$ to retrieve a set of centroid points $C = \{\bm{c}_i\}$.
	\item Split every centroid point into an input vector $\bm{v}_{\bm{c}_i}$ and output vector $\bm{u}_{\bm{c}_i}$.
	\item Define for every couple $(\bm{v}_{\bm{c}_i}, \bm{u}_{\bm{c}_i})$ the fuzzy rule: IF $\bm{x}$ is $\bm{v}_{\bm{c}_i}$ THEN $\bm{y}$ is $\bm{u}_{\bm{c}_i}$.
\end{enumerate}
We construct a dataset $(I,U)$ by having the teacher model $T$ interact with the environment. At every time step, the state vector $\bm{s}$ generated by the environment and the Q-vector $\bm{q}$ generated by $T$ are saved as the input and output vectors respectively. We then apply the procedure above using a Gaussian Mixture Model on this constructed dataset. The main advantage of a GMM is the fact that it allows us to specify a number of clusters \textit{a priori}, and provides us with a covariance matrix for each cluster. By restricting this matrix to be diagonal, we retrieve an estimate for the width of the corresponding fuzzy sets.

\subsection{Distillation}
\label{sec:distillation}
We apply policy distillation as described in \ref{sec:policy_distillation} using an $\epsilon$-greedy student policy. The student model $S$ is initialized using the procedure described in \ref{sec:pre-processing}. We use the neuro-fuzzy controller using weighted T-norms discussed in section \ref{sec:neuro_fuzzy_control_using_weighted_tnorms} with constant output functions: $f_j(\bm{x}) = \bm{y}_j$. This allows us to easily interpret the consequents of fuzzy rules as actions, namely the action corresponding to the highest Q-value in $\bm{y}_j$. As loss function, we use the temperature KL-divergence (Eq. \ref{eqn:temp-kl-div}) with temperature $\tau = 0.1$. The complete algorithm is described in Algorithm \ref{algo:distillation-base}.
\begin{figure}
	\hspace*{\algorithmicindent} \textbf{Input} $T$: Teacher (DQN) \\
	\hspace*{\algorithmicindent} \textbf{Output} $S$: Student (neuro-fuzzy controller)
	\begin{algorithmic}[1]
		\State Initialize replay memory $D$ of size $N$
		\ForAll{episodes}
		\State Get initial state vector $\bm{s}$
		\While{episode not done}
		\State Select action $a$ using $\epsilon$-greedy student policy
		\State Execute action $a$, receive next state $\bm{s}'$
		\State $\bm{q}^T \gets T(\bm{s})$
		\State Save $(\bm{s}, \bm{q}^T)$ in $D$
		\State $\bm{s} \gets \bm{s}'$
		\State Sample minibatch $D' \gets \{(\bm{s}_i,\bm{q}^T_i)\}$ from $D$
		\State Perform an optimization step on $S$ using $D'$
		\EndWhile
		\EndFor
	\end{algorithmic}
	\caption{Distillation algorithm}\label{algo:distillation-base}
\end{figure}
\subsection{Post-Processing}
\label{sec:post-processing}
The distillation algorithm tends to create very similar fuzzy sets in the rule antecedent parts. This leads to a lack of transparency in the rule base. We mitigate this problem by merging similar fuzzy sets. We do this by calculating the Jaccard index between fuzzy sets and iteratively merging the most similar fuzzy sets, as long as the similarity between these sets is greater than a certain threshold value. This threshold value is chosen as a hyperparameter, in this work we use $\alpha = 0.95$. The new, merged fuzzy set is then again a candidate for merging in the next iterations. Two sets $A_{ik}$ and $A_{il}$ (the fuzzy sets for input dimension $i$ in fuzzy rule $k$ and $l$, respectively) are merged as follows \cite{Paiva2004}:
\begin{align*}
\bar{\mu} &= \frac{n_{ik}\mu_{ik} + n_{il}\mu_{il}}{n_{ik}+n_{il}}\\
\bar{\sigma} &= \frac{n_{ik}\sigma_{ik} + n_{il}\sigma_{il}}{n_{ik}+n_{il}}
\end{align*}
Where $n_{ik}$ and $n_{il}$ are the number of sets that have been merged to create $A_{ik}$ and $A_{il}$ respectively. This formula assigns more weight to sets that are the result of many merges, so that the result of $n$ merges is the average of the values for the $n+1$ merged sets. The full algorithm is described in Algorithm \ref{algo:merge-sets}.

Since we use weighted T-norms, every rule has an importance weight associated to every input dimension. We simplify the rule base further in a second post-processing step by removing terms in the antecedents with importance weight smaller than $0.01$. This allows us to only consider the actually important input dimensions for every rule.

\begin{figure}
	\begin{algorithmic}[1]
		\State $n_{ij} \gets 1, \forall$ input dimension $i$, fuzzy rule $j$
		\ForAll{Input dimension $i$}
		\While{$\max\limits_{k \neq l}(J(A_{ik},A_{il})) > \alpha$}
		\State $(k,l) \gets \arg \max\limits_{k \neq l}(J(A_{ik},A_{il}))$
		\State $\mu' \gets \frac{n_{ik}\mu_{ik} + n_{il}\mu_{il}}{n_{ik}+n_{il}}$
		\State $\sigma' \gets \frac{n_{ik}\sigma_{ik} + n_{il}\sigma_{il}}{n_{ik}+n_{il}}$
		\State $A'(x) \gets \exp(-(\frac{x - \mu'}{\sigma'})^2)$
		\State Replace $A_{ik}$, $A_{il}$ with $A'$
		\EndWhile
		\EndFor
	\end{algorithmic}
	\caption{Antecedent set merging}\label{algo:merge-sets}
	\vspace{-1.5em}
\end{figure}

\subsection{Regularization}
\label{sec:regularization}
In order to produce fuzzy rule bases that can easily be simplified using the post-processing techniques described in section \ref{sec:post-processing}, we introduce two regularization terms during the distillation process. The first term encourages sets with similar $\mu$ values to grow towards each other, so they can easily be merged after training:
$$
\mathcal{L}_{merge} = \sum_{i=1}^{m} \sum_{k=1}^{n} \sum_{l=1}^{n} \frac{\sqrt{(\sigma_{ik} - \sigma_{il})^2}}{1 + \sqrt{(\mu_{ik} - \mu_{il})^2}}
$$
The first sum iterates over the $m$ input dimensions, while the second and third sums each iterate over the $n$ fuzzy rules. This regularization term is large if the distance between two centers is small, but the difference in standard deviation of the sets is large. This indicates strongly overlapping sets that cannot be merged easily. Such pairs of sets are also often difficult to interpret.

The second term is an L1 regularization on the T-norm importance weights. We first normalize the weights by dividing each weight by the largest importance weight for the corresponding rule. This prevents the weights from taking arbitrarily small values, as the same normalization is applied before calculation of the weighted T-norm (see section \ref{sec:neuro_fuzzy_control_using_weighted_tnorms}). The first sum again iterates over the $m$ input dimensions, and the second sum iterates over the $n$ fuzzy rules.
\begin{equation}
\mathcal{L}_{tnorm} = \sum_{i=1}^m \sum_{j=1}^n \frac{w_{ij}}{\max\limits_{k \in \{1, \dots, n\}} w_{ik}}
\end{equation}
The complete loss can be written as:
\begin{equation}
\mathcal{L} = \mathcal{L}_{KL}(\bm{Q}^T, \bm{Q}^S) + \lambda_{m} \mathcal{L}_{merge} + \lambda_{t} \mathcal{L}_{tnorm}
\end{equation}
Where $\bm{Q}^T$ and $\bm{Q}^S$ are batches of 64 teacher and student output vectors respectively. $\lambda_{m}$ and $\lambda_{t}$ are hyperparameters that indicate the weight of the regularization terms. In this work, we use $\lambda_{m} = 1$ and $\lambda_{t} = 0.5$.

\section{Results}
\label{sec:results}
\begin{figure}
	\centering
	\subfloat[CartPole]{\includegraphics[width=0.49\linewidth]{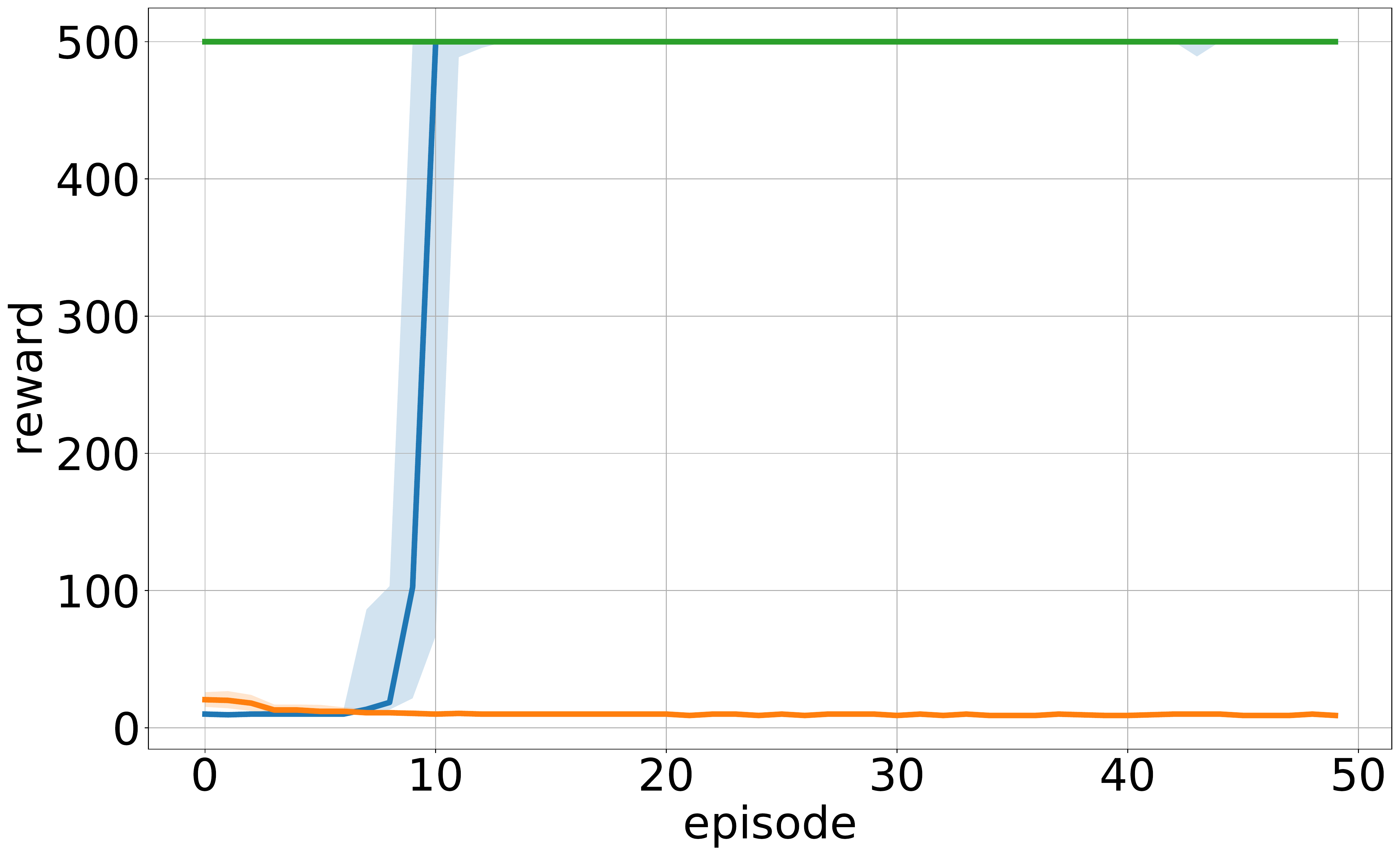}\label{fig:distill-results:CP}}
	\subfloat[MountainCar]{\includegraphics[width=0.49\linewidth]{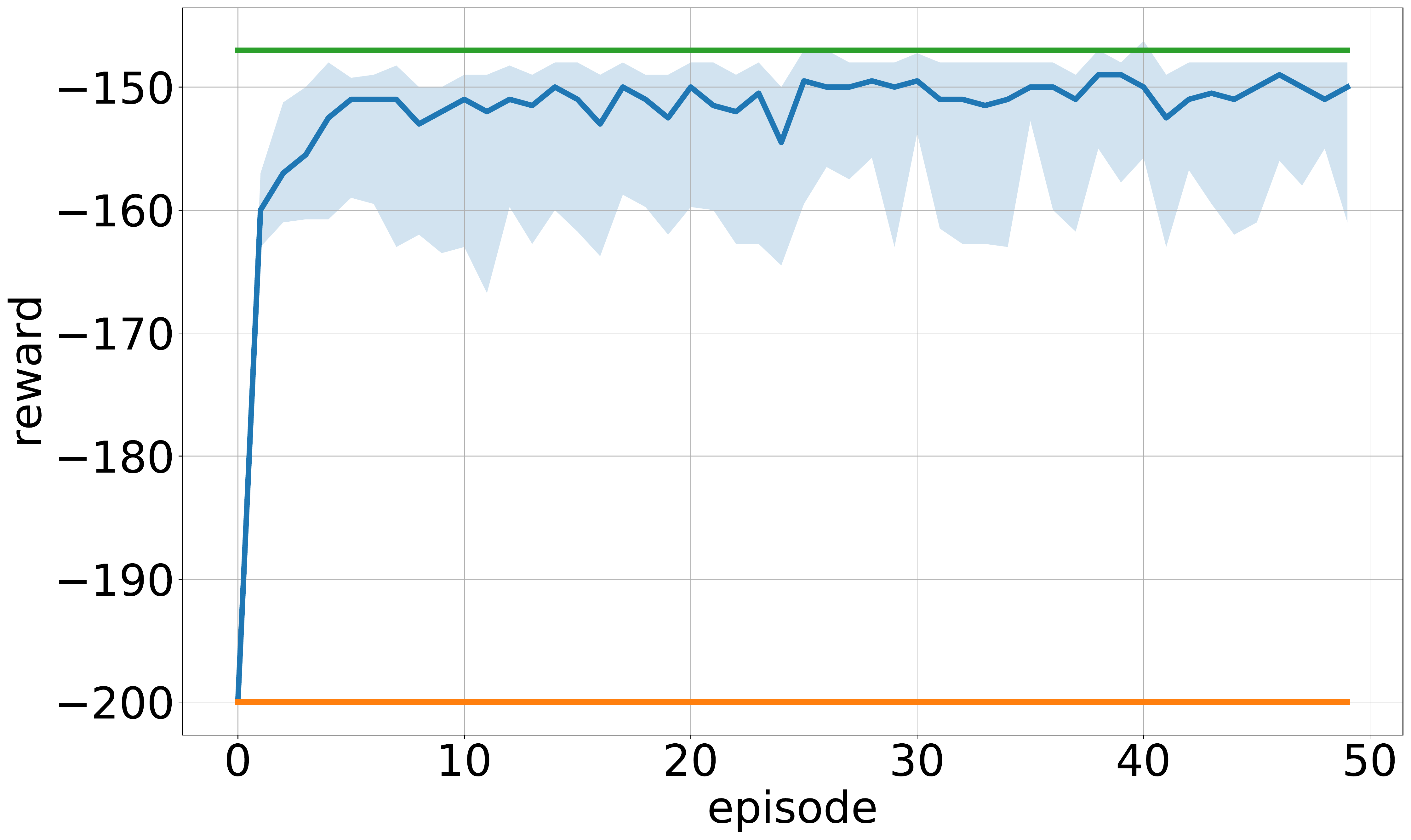}\label{fig:distill-results:MC}}
	\hfil
	\subfloat[LunarLander (4 rules)]{\includegraphics[width=0.49\linewidth]{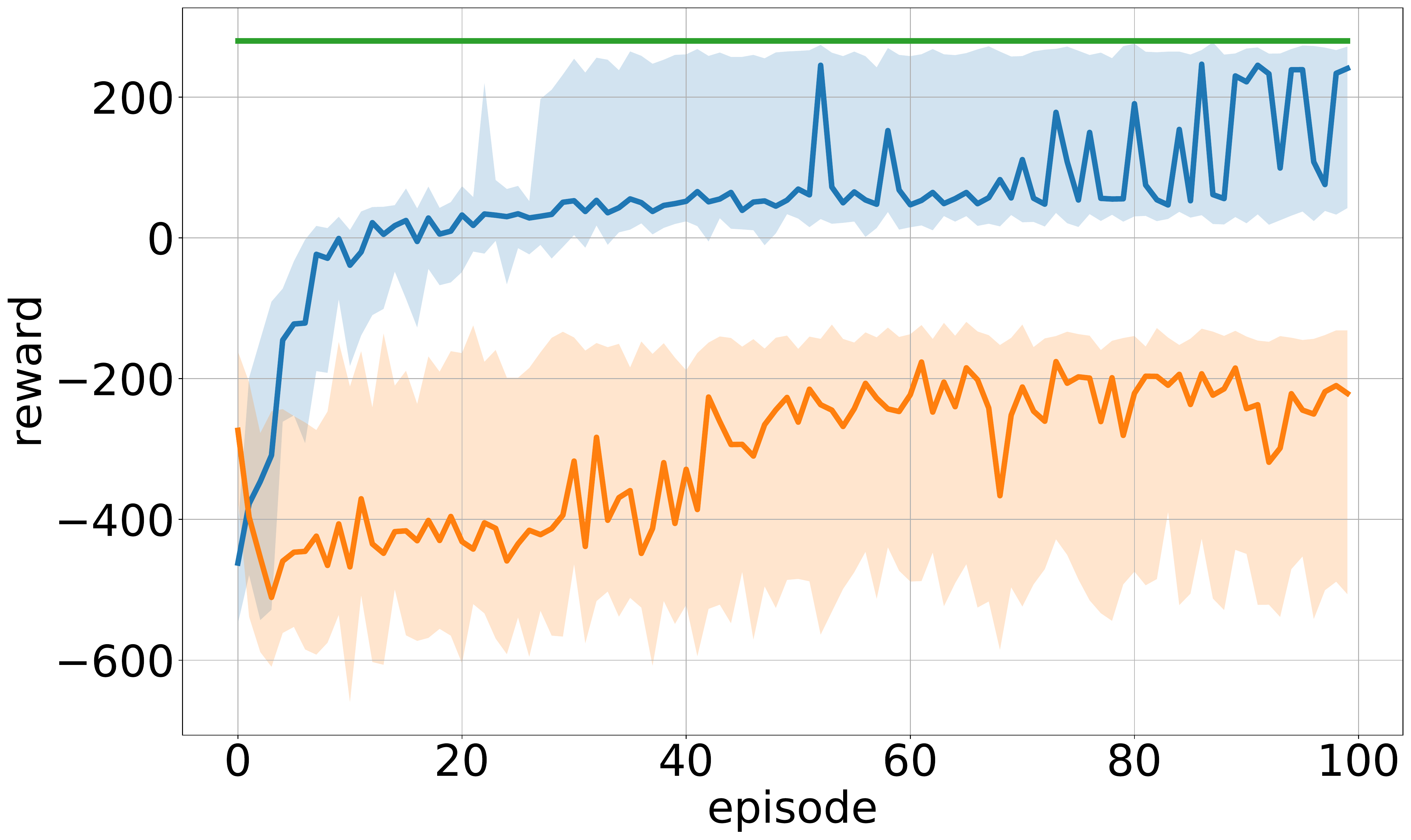}\label{fig:distill-results:LL4}}
	\subfloat[LunarLander (6 rules)]{\includegraphics[width=0.49\linewidth]{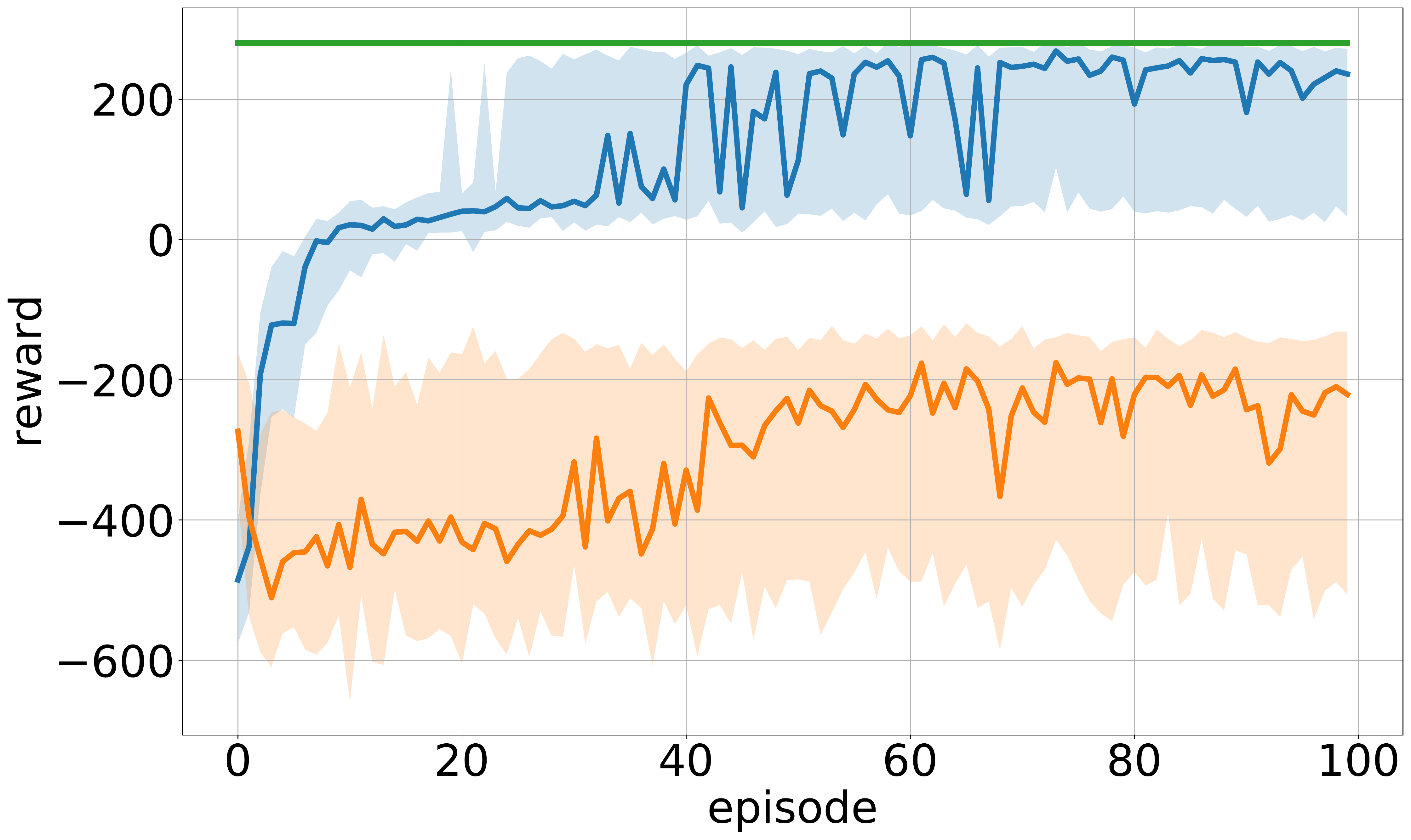}\label{fig:distill-results:LL6}}
	\caption{Distillation results. We ran the distillation process 50 times for each environment. Each figure shows the median, first quantile and third quantile of the achieved reward for a given environment using the distillation algorithm (blue curve) and using naive substitution of the deep Q-network with a neuro-fuzzy controller (orange curve). The green line indicates the median reward achieved by the teacher agent over 50 episodes.}
	\label{fig:distill-results}
	\vspace{-1.5em}
\end{figure}

We apply both the distillation algorithm and naive substitution in DQN (in which we simply substitute the deep neural network in DQN with a neuro-fuzzy controller) to three different OpenAI Gym \cite{Brockman2016} environments: CartPole, MountainCar and LunarLander. In the CartPole environment, the agent is tasked to balance a pole attached to a cart by applying a force of +1 or -1 to the cart. In the MountainCar environment, an underpowered car on a one-dimensional track is positioned between two hills. The goal is to drive the car up the right hill, by driving back and forth to build momentum. Finally, in the LunarLander environment the agent controls a 2D space ship and needs to land it on a landing platform. The results are summarized in Figure \ref{fig:distill-results}. In all three cases, the teacher model was a deep Q-network with 64 hidden nodes followed by a BatchNorm layer\footnote{Models retrieved from \url{https://github.com/araffin/rl-baselines-zoo}}.

In all three environments, we see that naive substitution of the deep neural network with a neuro-fuzzy controller is unable to learn a satisfying policy. Figure \ref{fig:distill-results} shows results obtained using the same number of rules as used in the distillation algorithm. Experiments with double or triple the number of rules had similar results, even when run for more than three times the number of episodes.

In the CartPole and MountainCar environments, we used two fuzzy rules to encode the student policy. This is the absolute minimum of rules, as a single fuzzy rule would effectively encode a constant policy. The CartPole environment has two available actions (push left, push right), so the two rules can be interpreted as ``when to move the cart left/right''. Remarkably, the MountainCar environment has three actions (push left, no push, push right). This means the fuzzy policy only uses two of the three available actions, never using the ``no push'' action. In both environments, we see very fast convergence, after 10 episodes for CartPole and only 5 episodes for MountainCar, although MountainCar has more variance in the reward which can mostly be attributed to variance in starting conditions.

Lunarlander is the most challenging environment. It has 8 input dimensions, two of which (ground left and ground right) take only two discrete values (1 or 0). This causes the system to define degenerate fuzzy sets, which negatively affects both performance and interpretability. This is a limitation of the distillation algorithm, and finding intelligent ways to incorporate discrete inputs in neuro-fuzzy systems is a possible direction for future research. In this case, the best results were achieved by simply ignoring the two discrete inputs. We tested the algorithm with both 4 or 6 fuzzy rules. With 4 fuzzy rules (which is the number of available actions), the system is able to match teacher performance in about half of the cases. More consistent performance is reached with 6 rules, at the cost of system simplicity.

We now give some examples of the resulting neuro-fuzzy controllers created by the distillation algorithm. The fuzzy sets are shown in graphs, combined with a table describing the accompanying fuzzy rules (see Figures \ref{fig:agent-cartpole}, \ref{fig:agent-mountaincar}, and \ref{fig:agent-lunarlander}). The opacity of the background color in the table shows the importance of the set in the corresponding fuzzy rule. For example, the first row of the table in Figure \ref{fig:agent-cartpole} should be read as: ``IF position IS A0\_0 with importance 0.07, AND angle IS A2\_0 with importance 0.07, AND velocity at tip is A3\_0 with importance 1.0, THEN the output is (2.88, -1.70)''. The outputs are the Q-values for the available actions, so this rule pushes the cart to the left.
\subsection{CartPole}
\begin{figure}
	\centering
	\subfloat{\includegraphics[width=\columnwidth]{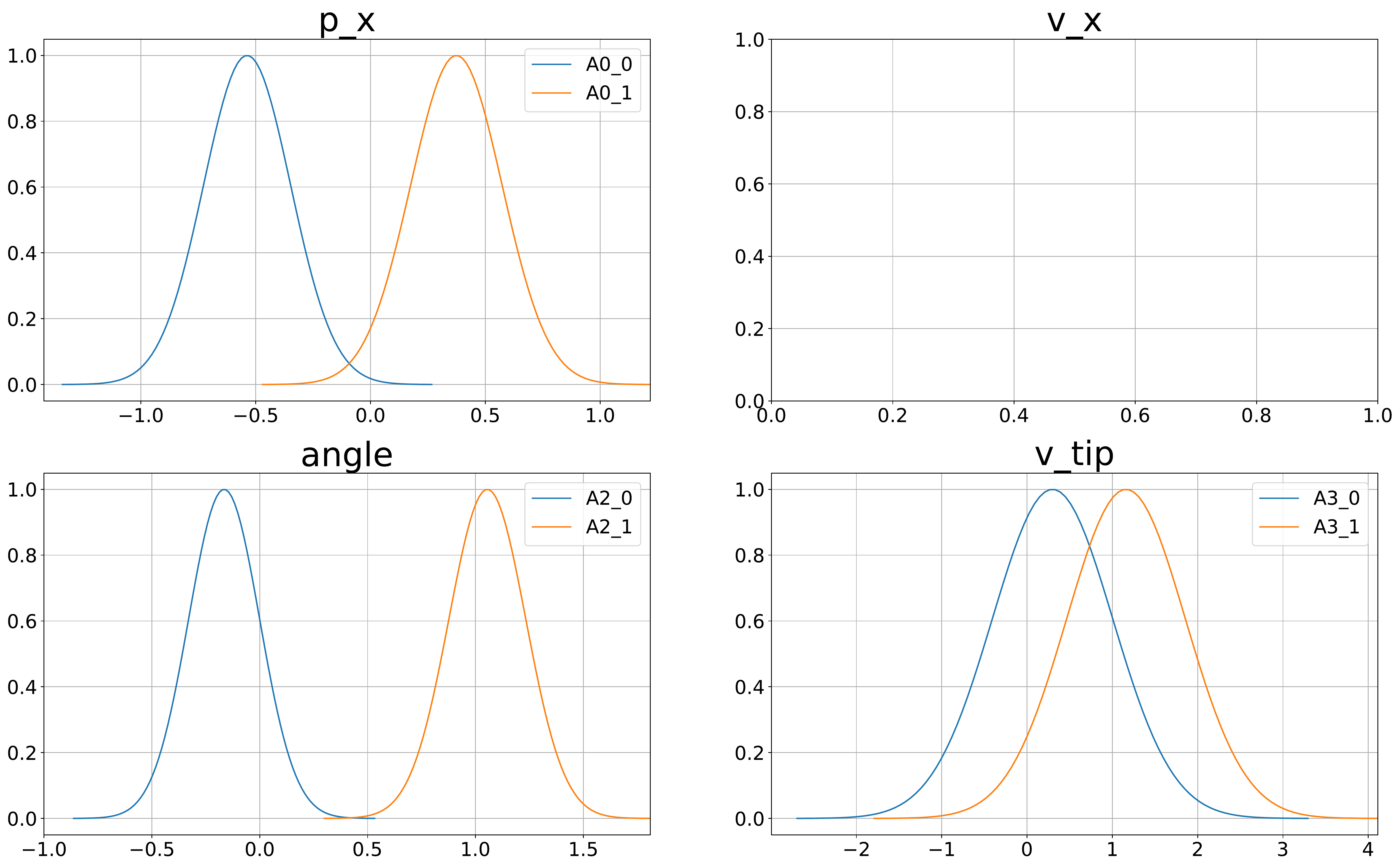}}
	\hfil
	\subfloat{
		\begin{tabular}{| c | c | c || c | }
			\hline
			{\tt p\_x} & {\tt angle} & {\tt v\_tip} & output \\
			\hline
			\cellcolor{red!7}A0\_0 & \cellcolor{red!7}A2\_0 & \cellcolor{red!100}A3\_0 & (2.88, -1.70) \\ \hline
			\cellcolor{red!2}A0\_1 & \cellcolor{red!4}A2\_1 & \cellcolor{red!100}A3\_1 & (-0.01, 1.18) \\ \hline
		\end{tabular}
	}
	\caption{\label{fig:agent-cartpole}Fuzzy sets (top) and rules (bottom) for the distilled CartPole agent. All fuzzy sets for \texttt{v\_x} were removed by the post-processing step, effectively removing this input. Intensity of the background color (bottom) is proportional to the importance weight of the corresponding fuzzy set.}
\end{figure}

Figure \ref{fig:agent-cartpole} shows a resulting neuro-fuzzy controller for the CartPole environment. We can see that every input dimension has clearly distinguishable ``low'' and ``high'' sets. The algorithm also assigned very low importance ($<0.01$) to each of the fuzzy sets in the second input dimension, which caused them to be masked away in the post-processing step. This means that the controller effectively learned that this input dimension is not necessary to complete the task. We also see a very sharp contrast in importance values of the rules, with almost all importance assigned to the {\tt v\_tip} dimension (velocity at the tip of the pole). We can conclude from this that the velocity at the tip of the pole is considered as much more important than the other dimensions by this agent.

\subsection{MountainCar}
\begin{figure}
	\centering
	\subfloat{\includegraphics[width=\columnwidth]{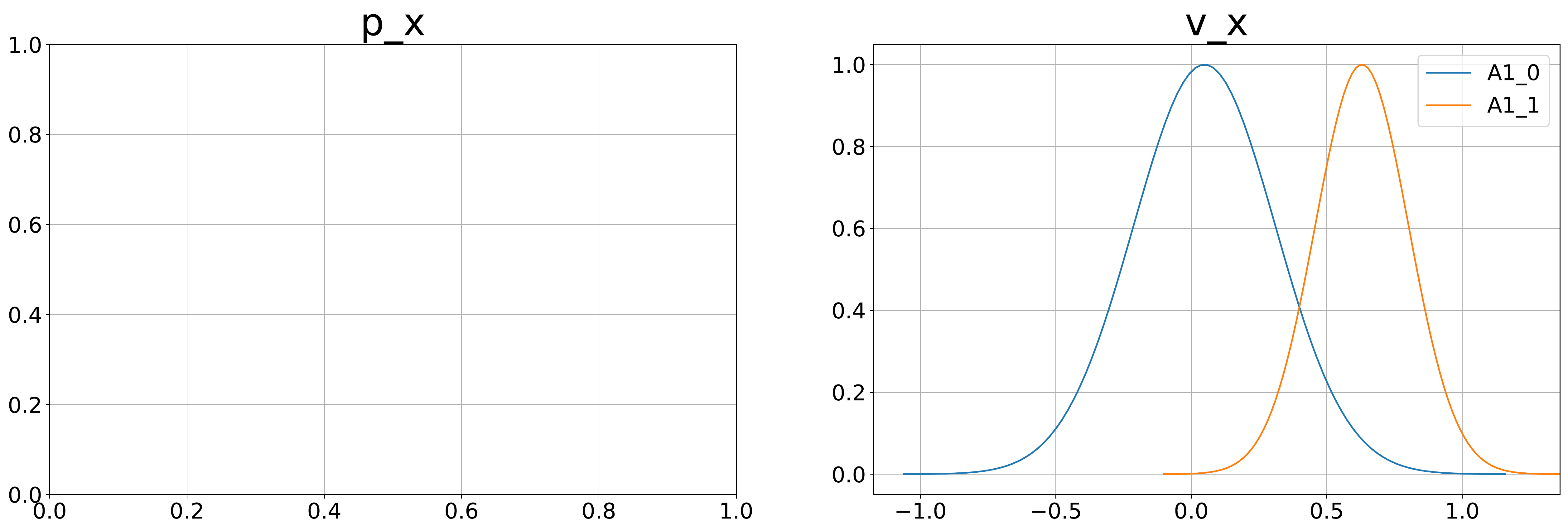}}
	\hfil
	\subfloat{
		\begin{tabular}{| c || c | }
			\hline
			{\tt v\_x} & output \\
			\hline
			A1\_0 & (4.61, 0.47, -9.53) \\ \hline
			A1\_1 & (-0.42, -0.50, 2.62) \\ \hline
		\end{tabular}	
	}
	\caption{\label{fig:agent-mountaincar}Fuzzy sets (top) and rules (bottom) for the distilled MountainCar agent. All fuzzy sets for \texttt{p\_x} were removed by the post-processing step, meaning that the controller is only influenced by \texttt{v\_x}.}
\end{figure}

Figure \ref{fig:agent-mountaincar} shows a resulting controller on the MountainCar environment. We can see very similar results to those obtained with CartPole. In this case, the algorithm again ignores one of the two input dimensions, reaching the same performance as the teacher using only the velocity of the car. This makes sense, as the optimal policy can be viewed as simply ``amplifying'' the current velocity (going left to build up potential energy or going right to try and reach the goal) until the goal is reached.

\subsection{LunarLander}
\begin{figure*}
	\centering
	\subfloat{\includegraphics[width=\linewidth]{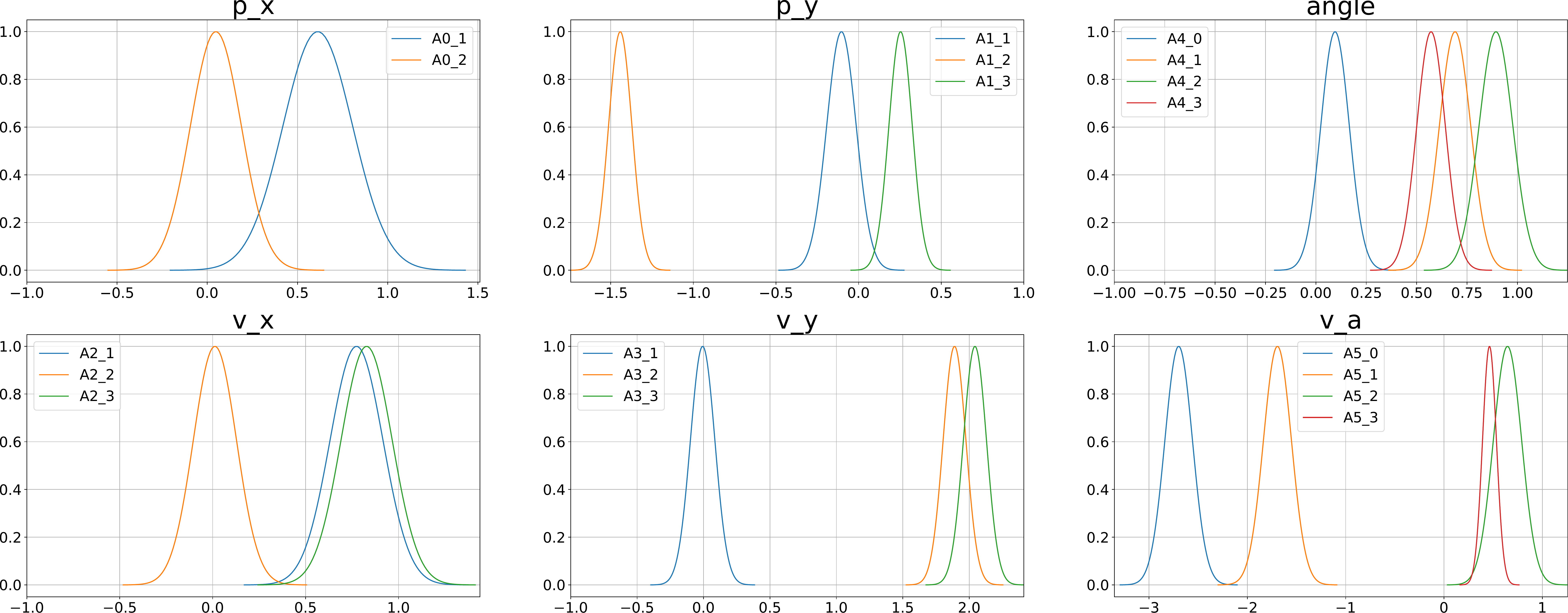}}
	\hfil
	\subfloat{
		\begin{tabular}{| c | c | c | c | c | c || c | }
			\hline
			{\tt p\_x} & {\tt p\_y} & {\tt v\_x} & {\tt v\_y} & {\tt angle} & {\tt v\_a} & output \\
			\hline
			- & - & - & - & \cellcolor{red!13}A4\_0 & \cellcolor{red!100}A5\_0 & (-3.10, 2.21, 1.34, -2.40) \\ \hline
			\cellcolor{red!5}A0\_1 & \cellcolor{red!3}A1\_1 & \cellcolor{red!9}A2\_1 & \cellcolor{red!12}A3\_1 & \cellcolor{red!9}A4\_1 & \cellcolor{red!100}A5\_2 & (-8.90, -0.63, 5.04, -2.37) \\ \hline
			\cellcolor{red!8}A0\_2 & \cellcolor{red!47}A1\_2 & \cellcolor{red!16}A2\_2 & \cellcolor{red!47}A3\_2 & \cellcolor{red!12}A4\_2 & \cellcolor{red!100}A5\_2 & (0.82, -6.59, -0.45, 3.80) \\ \hline
			- & \cellcolor{red!5}A1\_3 & \cellcolor{red!8}A2\_3 & \cellcolor{red!100}A3\_3 & \cellcolor{red!10}A4\_3 & \cellcolor{red!7}A5\_3 & (2.84, 4.15, -3.03, -2.93) \\ \hline
		\end{tabular}
	}
	\caption{\label{fig:agent-lunarlander}Fuzzy sets (top) and rules (bottom) for the distilled LunarLander agent. Intensity of the background color (bottom) is proportional to the importance weight of the corresponding fuzzy set.}
\end{figure*}

Finally, we show a 4-rule neuro-fuzzy controller that matches teacher performance on the LunarLander environment in Figure \ref{fig:agent-lunarlander}. The input now consists of 8 dimensions, but the last two ({\tt gnd\_left} and {\tt gnd\_right}) are discarded as they only take two discrete values (1 or 0). Incorporating those discrete features causes the system to define degenerate fuzzy sets, which negatively affects both performance and interpretability, a limitation of the algorithm. The rules are now not as easily interpretable as in the CartPole and MountainCar environments because of the relatively large number of input dimensions, but fuzzy partitions are still mostly interpretable and the system is able to identify a subset of input dimensions that is significantly more important than the others, namely {\tt v\_y} (vertical velocity) and {\tt v\_a} (angular velocity). Although we cannot easily and intuitively summarize the entire policy, this still brings some insight into the inner workings of the system.

\section{Discussion}
\label{sec:discussion}
Although this technique is feasible for certain environments, there are some obvious limitations to the neuro-fuzzy architecture. First of all, it only makes sense to use this architecture if input features carry a semantic meaning. If the input consists of raw pixel values for example, we cannot reasonably expect fuzzy rules defined on these inputs to be interpretable. A possible direction for future work could be to explore techniques for extracting disentangled representations (e.g. using Variational Auto-Encoders \cite{Mathieu}) and to use these representations as input features for a neuro-fuzzy controller.

In conclusion, we have demonstrated how distillation can be used to train compact neuro-fuzzy controllers to solve tasks that they are unable to solve directly. This technique opens the path to a new way of creating fuzzy controllers: instead of manually designing a compact fuzzy controller, or training a (much less compact) neuro-fuzzy controller, we can first train a neural network to solve a task using Q-learning and then automatically distill this policy into a compact neuro-fuzzy controller. Although this approach was demonstrated only on toy environments in this work, it could be applied to real-world control systems such as robotic control \cite{chatterjee2005} or non-linear modeling of dynamic systems \cite{bartczuk2016}. Finally, more future work can be done to combine this technique with existing literature in the field of (neuro-)fuzzy control. This includes using alternative membership functions, training techniques \cite{Hein2017} or techniques for rule base simplification \cite{Setnes1998}.

\bibliographystyle{IEEEtran}
\bibliography{bibliography}

\end{document}